\documentclass[letterpaper,journal]{IEEEtran}

\usepackage{amsmath,amssymb,amsfonts}
\usepackage{booktabs}
\usepackage{graphicx}
\usepackage{tabularx}
\usepackage{array}
\usepackage{stfloats}
\usepackage{url}
\usepackage{xcolor}
\usepackage{dsfont}
\usepackage{multirow} 
\usepackage{colortbl}
\usepackage{cite}

\title{Lean-SAM2: Target-Anchored Memory and Encoder \\ Acceleration for SAM2}

\author{Xudong Ouyang, Wenlun Zhang, Yimin Xu, Huazhong Liu, Yunshan Zhong
\thanks{
X. Ouyang is with the School of Biomedical Engineering, Hainan University, Hainan 570228, China.

W. Zhang is with the Department of Electronics and Electrical Engineering, Keio University, Kanagawa 223-8522, Japan.

Y. Xu is with the College of Computer Science, Shenyang Aerospace University, Shenyang 110135, China.

H. Liu and Y. Zhong (Corresponding  Author) are with the School of Computer Science and Technology, Hainan University, Hainan 570228, China (e-mail: yszhong01@gmail.com).}
}

\begin{document}
\maketitle

\begin{abstract}

The Segment Anything Model 2 (SAM2) has advanced temporal promptable segmentation, yet its deployment remains hindered by heavy memory cross-attention overhead and redundant full-frame visual feature extraction. 
While recent methods explore efficiency via heuristic memory pruning and window-based sparse routing, they typically suffer from catastrophic performance degradation in complex segmentation scenarios replete with occlusions and distractors. 
To resolve these limitations, we propose \textbf{Lean-SAM2}, a holistic lightweight framework designed to address the above vulnerabilities while systematically eliminating computational redundancies. 
Specifically, Lean-SAM2 integrates three collaborative mechanisms: 
(1) \textbf{Target-Anchored Memory Pruning (TAMP)} safeguards target tokens against deceptive attention by modulating raw attention significance with semantic consistency against prompt-derived foreground anchors; 
(2) \textbf{Temporal Condensation with Insurance Memory (TCIM)} condenses historical context via a visibility-gated fusion while conditionally archiving high-confidence entries in a parallel insurance bank; 
and (3) \textbf{Target-Anchored Risk-Aware Routing (TARR)} selectively activates the heavy image encoder for target-related windows based on anchor similarity, utilizing a risk-aware fallback policy to trigger full-frame refreshes during volatile transitions. 
Extensive evaluations across multiple challenging benchmarks demonstrate that Lean-SAM2 establishes a superior balance between accuracy and efficiency. For example, on the LVOSv2 validation dataset, Lean-SAM2 achieves overall inference speedups of $1.412\times$ and $1.417\times$ on the SAM2.1-Large and SAM2.1-Base+, respectively, significantly outperforming Efficient-SAM2 while boosting the corresponding $\mathcal{J}\&\mathcal{F}$ scores by $5.0\%$ and $3.6\%$. Code is available at \url{https://github.com/DeawhaleQwQ/Lean-SAM2}.

\end{abstract}


\section{Introduction}

\IEEEPARstart{T}{he} Segment Anything Model 2 (SAM2) \cite{ravi2025sam} has emerged as a landmark vision foundation model, extending the promptable segmentation capabilities of SAM~\cite{kirillov2023segment} into the temporal domain. 
By incorporating a continuous memory bank to store long-term historical features, SAM2 achieves remarkable performance in video object segmentation(VOS)~\cite{perazzi2016benchmark,cheng2021rethinking,cheng2022xmem,cheng2024putting}. 
Despite its exceptional performance, deploying SAM2 in real-time or resource-constrained scenarios remains heavily prohibited by two computational bottlenecks~\cite{zhang2026efficient,mandal2025fast,ding2026tinysam,liu2024surgical,xu2026tsms}. First, the token-level cross-attention footprint inflates due to a burdensome cached historical memory bank. Second, the visual backbone undergoes heavy full-frame feature extraction at every single time step, regardless of the target's actual dynamic information distribution volatility.

To alleviate these computational burdens, extensive efforts have been dedicated to investigating efficient SAM2 inference approaches~\cite{zhang2026efficient,xiong2025efficient,mandal2025fast,yang2024samurai,zhang2025ahcptq,farronato2025q,duan2026mix,ren2025q,tang2026efficient,zhou2025edgetam,liu2024surgical,xu2026tsms,zhang2026ahcqsamaccuratehardwarecompatibleposttraining}.  
Among these methods, post-training compression has received widespread attention due to its unique capacity to optimize inference efficiency with zero or little cost overhead. Along this line, a recent pioneering work~\cite{zhang2026efficient} proposes a lightweight post-training SAM2 acceleration method, incorporating heuristic memory pruning and image encoder sparse routing mechanisms. 
Specifically, the memory pruning mechanism reduces the memory bank computational overhead by discarding historical tokens that exhibit low cross-attention significance. 
Concurrently, the image encoder sparse routing mechanism mitigates computational overhead by partitioning incoming frames into windows and selectively routing target-related regions to the heavy image encoder based on mask predictions, while relegating background windows to a lightweight, trainable bypass network.

Although effective in clean videos with continuous target visibility, we reveal that this approach suffers from inherent vulnerabilities under complex conditions, such as severe occlusions and distractors. 
Specifically, as shown in Fig.~\ref{fig:tamp-corrupted-attention}, when semantically similar distractors temporarily corrupt attention scores, the attention-only memory pruning mechanism tends to discard indispensable target-related tokens, leading to irreversible information loss. 
Concurrently, as illustrated in Fig.~\ref{fig:case-study-routing}, if mask predictions drift under the influence of occlusions or distractors, the image encoder sparse routing mechanism mistakenly bypasses target-related regions, thereby triggering a vicious cycle of recursive feature degradation and catastrophic failure.

\begin{figure*}[htpb]
\centering
\includegraphics[width=\linewidth]{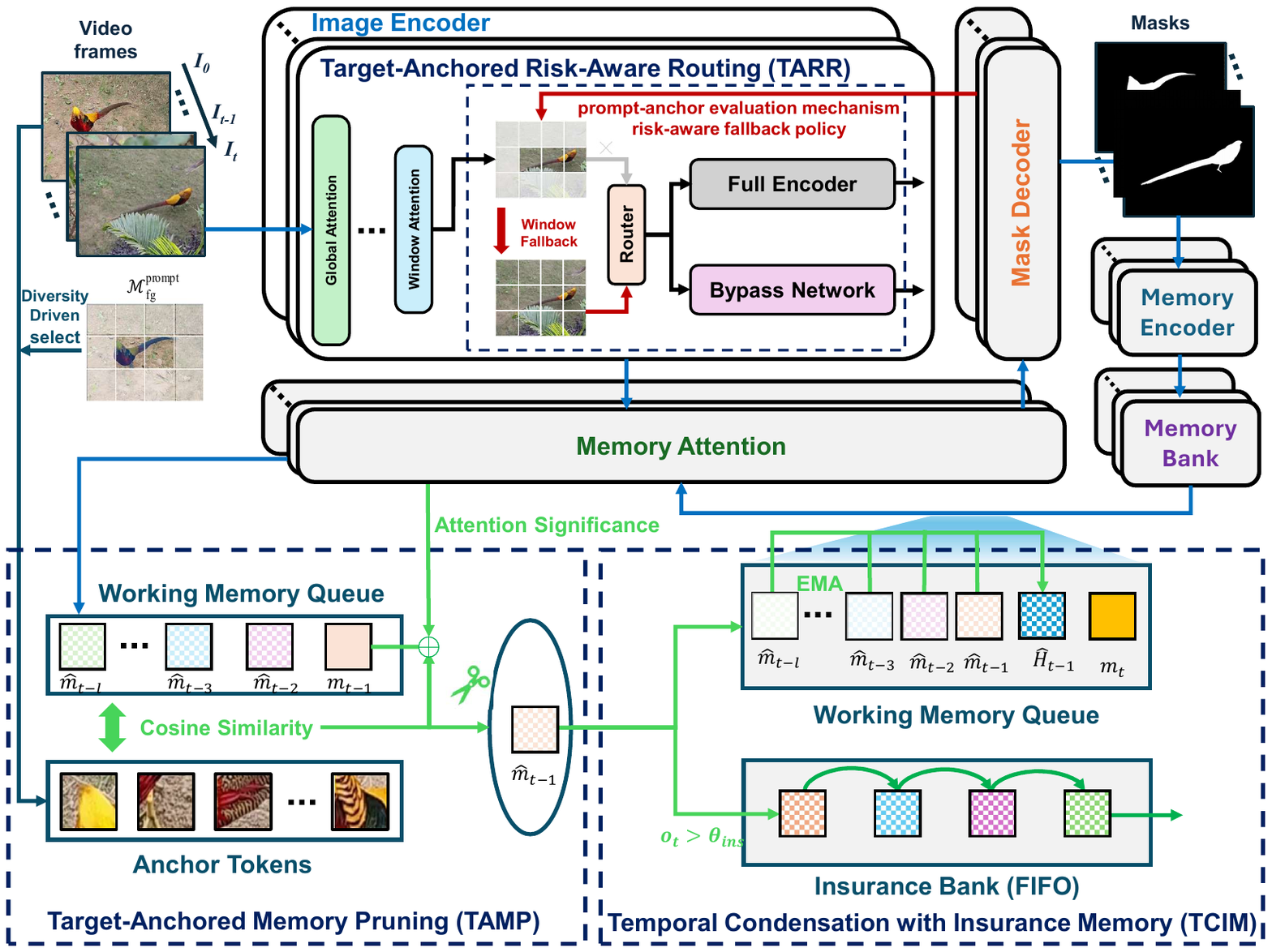}
\caption{The framework of the proposed Lean-SAM2. Lean-SAM2 introduces three collaborative mechanisms, including Target-Anchored Memory Pruning (TAMP), Temporal Condensation with Insurance Memory (TCIM), and Target-Anchored Risk-Aware Routing (TARR).}
\label{fig:framework}
\end{figure*}

In this paper, as shown in Fig.~\ref{fig:framework}, we introduce \textbf{Lean-SAM2}, a holistic framework designed to resolve the above vulnerabilities while systematically eliminating computational redundancy of SAM2. 
As illustrated in Fig.~\ref{fig:framework}, Lean-SAM2 incorporates three collaborative mechanisms, including Target-Anchored Memory Pruning (TAMP), Temporal Condensation with Insurance Memory (TCIM), and Target-Anchored Risk-Aware Routing (TARR).
First, TAMP mitigates intra-frame spatial redundancy within the memory bank. By modulating raw attention significance with anchor semantic consistency against a diverse set of prompt-derived foreground anchors, it steadily safeguards target-related tokens from being erroneously discarded even when attention is misguided by visual distractors.
Then, TCIM further alleviates inter-frame temporal redundancy within the memory bank by recursively condensing historical memory entries into a highly compact working queue via a target visibility-gated fusion. Concurrently, it maintains a parallel insurance bank to conditionally archive high-confidence entries, thereby preventing the loss of critical information during prolonged segmentation.
Finally, TARR protects sparse routing from prediction drift while still ensuring architectural efficiency. Guided by prompt-anchored similarity, it selectively routes target-aligned spatial windows to the full encoder. Concurrently, it employs a risk-aware fallback policy to trigger a full-frame refresh, thereby eliminating potential misguidance caused by occlusions and distractors.

Through the integration of these three mechanisms, Lean-SAM2 effectively accelerates SAM2 while preserving its strong performance. Extensive evaluations on five challenging VOS benchmarks demonstrate that Lean-SAM2 establishes a superior trade-off between inference speed and accuracy compared to existing methods. For instance, on the LVOSv2 validation dataset, Lean-SAM2 achieves overall inference speedups of $1.412\times$ and $1.417\times$ on the SAM2.1-Large and SAM2.1-Base+, respectively, significantly outperforming Efficient-SAM2 while boosting the corresponding $\mathcal{J}\&\mathcal{F}$ scores by $5.0\%$ and $3.6\%$.

\section{Related Work}

\subsection{Segment Anything Models} 
Segment Anything Model 2 (SAM2){\cite{ravi2025sam}} has extended the powerful promptable image segmentation paradigm of SAM{\cite{kirillov2023segment}} into the temporal domain by introducing a streaming memory mechanism. SAM2 propagates target information across video frames in a recurrent manner, extracting visual features with a hierarchical image encoder and retrieving historical contexts via memory attention. It achieves exceptional generalization performance in various video object segmentation (VOS) tasks~\cite{hong2025lvos,alansari2026sentry,perazzi2016benchmark,yilmaz2006object,zhou2022survey}
Beyond natural videos, SAM2 has been extensively adapted to the medical domain ~\cite{zhu2024medical,liu2024surgical,ma2025medsam2,xu2026tsms}, demonstrating remarkable generalization capabilities. Concurrently, recent advancements have extended SAM2 to long video \cite{ding2025sam2long} and highly complex scenarios. For instance, dedicated variants have been developed to tackle target drift and identity switches induced by fast-moving or self-occluding objects~\cite{yang2024samurai}, as well as to mitigate the interference of background distractors through advanced memory management~\cite{videnovic2025distractor}.
However, deploying SAM2 in practical application~\cite{hong2025lvos} remains heavily prohibited by its substantial computational cost. 
Specifically, memory attention and heavy image encoder constitute the primary efficiency bottlenecks of SAM2. Mitigating these computational redundancies is therefore crucial for efficient inference.

\subsection{Lightweight SAM2 and Post-training Acceleration}

Before the emergence of SAM2, a series of efficient SAM variants explored lightweight promptable image segmentation by replacing heavy image encoders, distilling compact models, quantizing models' weights and activation, or pruning redundant structures and tokens~\cite{zhao2023fast,zhang2023faster,xiong2024efficientsam,zhou2023edgesam,shu2025tinysam,chen2024slimsam,abebe2025supersam,tran2026sparsesam,liu2026structsam,wen2026car,zhang2026saq,zhang2025ahcptq,zhang2024efficientvit,lv2024ptq4sam,sun2025efficient}. These methods reveal the substantial redundancy in segmentation foundation models, but most of them focus on image-level SAM and cannot handle the unique structure of SAM2. 
 
Existing efficient SAM2 methods primarily reduce computation in two ways. 
The first line of work redesigns the network structure. 
EfficientTAM\cite{xiong2025efficient} revisits a lightweight plain vision encoder and efficient memory cross-attention.
EdgeTAM \cite{zhou2025edgetam} targets on-device deployment with a compact architecture and a 2D spatial perceiver for memory compression.
TinySAM2~\cite{ding2026tinysam} further combines a lightweight RepViT encoder~\cite{wang2024repvit} with memory quality management, spatial memory pooling, and temporal token selection to build an extremely compact model.
In addition, several parameter-efficient or domain-oriented SAM2 variants reduce training or deployment cost in medical and other resource-constrained scenarios through lightweight adapters, LoRA modules, prompt-free adaptation, or memory-aware fine-tuning ~\cite{mandal2025sam2lora,li2026uniultra,mitsuoka2026prompt,zhang2025sam2v,li2026mft}.
Although these methods have demonstrated substantial success, they typically rely on task-specific fine-tuning, which limits their flexibility compared with lightweight post-training acceleration methods.

In contrast, the second line of work explores post-training acceleration techniques such as quantization and pruning, which circumvent end-to-end retraining and thus offer greater flexibility and generalization potential for SAM2 adaptation.
Quantization-based methods such as Q-SAM2, Mix-QSAM2, Q-MiniSAM2, and CAR-SAM reduce the deployment cost of SAM2 via low-bit quantization or mixed-precision quantization~\cite{ren2025q,farronato2025q,duan2026mix,zhang2026ahcqsamaccuratehardwarecompatibleposttraining}
For instance, Mix-QSAM2~\cite{duan2026mix} combines importance-driven mixed-precision quantization with selective memory synthesis, assigning higher bit-widths to sensitive layers and compressing redundant historical frames in the memory bank. Q-MiniSAM2~\cite{ren2025q} reduces redundant memory and computation in attention through hierarchy-based video quantization and adaptive mutual-KV.

However, quantization-based methods typically require specialized hardware or low-bit operators to achieve actual speedups, whereas pruning-based methods can accelerate models on general devices.
Generic token pruning and merging methods reduce redundant tokens in Vision Transformers by discarding unimportant tokens or merging similar ones ~\cite{rao2021dynamicvit,liang2022not,yin2022vit,bolya2022token,bolya2023token,norouzi2024algm,kienzle2024segformer++,dang2023efficient,dang2024beyond,miao2024region,lin2023prototypical,wang2022delving,cheng2021rethinking}.
Building on this idea, a series of SAM2 acceleration methods have been developed, including sparse window routing, sparse memory retrieval, text-driven token pruning, frame pruning, memory-splitting pruning, recurrent dynamic submodel routing, and memory-aware token compression \cite{zhang2026efficient,mandal2025fast,tang2026efficient}
For example, RDS~\cite{tang2026efficient} uses a prediction-aware router to dynamically activate frame-specific blocks according to previous predictions and current visual features.
Efficient-SAM2~\cite{zhang2026efficient} presents an overall acceleration framework, which introduces an object-aware post-training pruning method for SAM2. It proposes Sparse Window Routing (SWR) to bypass background windows into a lightweight branch, and Sparse Memory Retrieval (SMR) to restrict memory attention to salient tokens.
Nevertheless, it fails to account for complex scenarios replete with occlusions and distractors, thus suffering from unsatisfactory performance.

\section{Preliminaries}

\subsection{SAM2 Pipeline.}

SAM2~\cite{ravi2025sam} extends promptable image segmentation to videos via a memory module, establishing a recurrent read-decode-write pipeline. 
For an incoming frame $I_t$, the image encoder $\mathbf{E}_{\mathrm{img}}$ extracts multi-stage hierarchical features:
\begin{equation}
  \{F_t^{s_0}, F_t^{s_1}, F_t^{s_2}\} = \mathbf{E}_{\mathrm{img}}(I_t),
  \label{eq:prelim-image-encoder}
\end{equation}
where $F_t^{s_2}$ is the low-resolution image embedding, and $\{F_t^{s_0}, F_t^{s_1}\}$ are high-resolution features.
To incorporate temporal context, $F_t^{s_2}$ queries the memory bank $\mathcal{M}^{\text{bank}}_t$ via a memory attention module $\mathbf{A}_{\mathrm{mem}}$ to generate the memory-conditioned feature $F_t$:
\begin{equation}
  F_t = \mathbf{A}_{\mathrm{mem}}(F_t^{s_2}, \mathcal{M}^{\text{bank}}_t).
  \label{eq:prelim-memory-attn}
\end{equation}

Concurrently, the prompt encoder yields embeddings $P_t$ and the mask decoder $\mathbf{D}_{\text{mask}}$ then predicts the object mask $O_t$ and occlusion score $o_t$ based on $F_t, F_t^{s_0}, F_t^{s_1}, P_t$:
\begin{equation}
  (O_t, o_t) = \mathbf{D}_{\text{mask}}(F_t, F_t^{s_0}, F_t^{s_1}, P_t),
  \label{eq:prelim-decoder}
\end{equation}
where $O_t$ is the predicted mask output and $o_t$ denotes the occlusion score.
Finally, the memory encoder $\mathbf{E}_{\mathrm{mem}}$ fuses the current prediction with the image features to generate and store a new memory entry:
\begin{equation}
  m_t = \mathbf{E}_{\mathrm{mem}}(F_t^{s_2}, O_t), \quad m_t \in \mathbb{R}^{H \times W \times C},
  \label{eq:prelim-memory-write}
\end{equation}
where $H$ and $W$ are the spatial height and width of the memory entry.

\textbf{Image Encoder.} The hierarchical image encoder $\mathbf{E}_{\mathrm{img}}$ utilizes windowed self-attention~\cite{ravi2025sam} to capture local structures. For an input $X^{k-1} \in \mathbb{R}^{H \times W \times C}$ at layer $k$, the feature map is partitioned into non-overlapping local windows of size $(h, w)$ for Multi-Head Self-Attention (MHSA) before being unpartitioned:
\begin{align}
 & X^{k-1}_W = \mathrm{WindowPartition}(X^{k-1}; h, w), \\
  & \widehat{X}_W^{k}   = \mathrm{MHSA}(X_W^{k-1}), \\
  & X^{k} = \mathrm{WindowUnpartition}(\widehat{X}_W^{k}; h, w).
  \label{eq:prelim-window-attn}
\end{align}

\textbf{Memory Bank.} 
In semi-supervised VOS tasks, the memory bank $\mathcal{M}^{\text{bank}}_t$ aggregates the prompt identity cues $\mathcal{M}^{\mathrm{prompt}}$ (from the annotated initial frame) and a continuous working memory queue $\mathcal{M}_t$ of length $l$ containing recent frames under a First-In-First-Out (FIFO) policy:
\begin{equation}
\begin{aligned}
  \mathcal{M}^{\text{bank}}_t &= \mathcal{M}^{\mathrm{prompt}} \oplus \mathcal{M}_t, \,\,
  \mathcal{M}_t = \{m_{t-1}, \cdots, m_{t-l}\},
\end{aligned}
\label{eq:prelim-memory-bank}
\end{equation}
where $\oplus$ denotes token-dimension concatenation.

\textbf{Memory Attention.}
The memory attention module $\mathbf{A}_{\mathrm{mem}}$ performs cross-attention where $F_t^{s_2}$ acts as the Query, and the token sequence $\Phi(\mathcal{M}^{\text{bank}}_t)$ extracted from the memory bank provides the Keys and Values:
\begin{align}
  Q_t = F_t^{s_2} W_Q, \,\, K_t &= \Phi(\mathcal{M}^{\text{bank}}_t) W_K, \,\, V_t = \Phi(\mathcal{M}^{\text{bank}}_t) W_V, \\
  F_t = A_tV_t & = \mathrm{softmax}\left( \frac{Q_t K_t^{\top}}{\sqrt{d}} \right) V_t,
  \label{eq:prelim-qkv}
\end{align}
where $W_Q, W_K, W_V$ are projection matrices.


\subsection{Bottleneck in SAM2}

\begin{figure}[!thpt]
\centering
\includegraphics[width=0.8\linewidth]{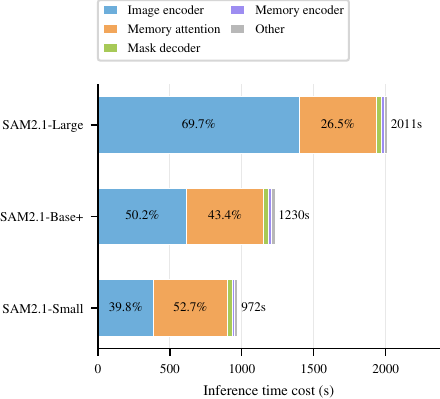}
\caption{Runtime latency decomposition of SAM2 across various model scales. We report the inference time for SAM2.1-Small, -Base+, and -Large backbones evaluated on 20 videos from the LVOSv2 valid dataset using a single NVIDIA RTX 3090 GPU. Notably, the image encoder and memory attention consistently dominate the computational cost.}
\label{fig:small-tiny-speed-runtime-breakdown}
\end{figure}

The memory attention and image encoder modules dominate the primary computational and memory footprints of SAM2 inference.
As shown in Fig.~\ref{fig:small-tiny-speed-runtime-breakdown}, the runtime analysis reveals that the memory attention and image encoder together consume the vast majority of the video segmentation time across all model scales, accounting for over 92\% of the inference latency on SAM2-Small, 83\% on SAM2-Base+, and up to 96\% on SAM2-Large.
In summary, these empirical observations motivate the need to jointly alleviate the dual computational and memory overheads localized within the image encoder and memory attention processes.

\section{Method}
\label{sec:method}

\subsection{Overview}
\label{sec:method-overview}

As shown in Fig.~\ref{fig:framework}, we propose Lean-SAM2, a unified framework that accelerates SAM2 via three complementary mechanisms including Target-Anchored Memory Pruning (TAMP), Temporal Condensation with Insurance Memory (TCIM), and Target-Anchored Risk-Aware Routing (TARR).
Specifically, Lean-SAM2 first introduces TAMP to alleviate intra-frame spatial redundancy within the memory bank. By integrating attention significance with semantic consistency against a compact set of diverse anchors sampled from the initial prompt foreground, TAMP safeguards target-related tokens from being erroneously discarded when attention becomes misguided by distractors. 
Subsequently, Lean-SAM2 presents TCIM to further mitigate inter-frame temporal redundancy within the memory bank. TCIM recursively compresses historical memory entries into a highly compact working queue based on a visibility-gated fusion, while concurrently maintaining a parallel insurance bank that conditionally archives high-confidence entries to preserve critical long-term context.
Finally, Lean-SAM2 introduces TARR to safeguard window-level image encoder sparse routing against prediction drift while sustaining architectural efficiency. Guided by prompt-anchored similarity, TARR selectively routes target-aligned spatial windows to the heavy image encoder. Concurrently, a risk-aware fallback policy triggers a full-frame refresh during ambiguous tracking phases, thereby eliminating potential misguidance caused by occlusions and distractors.

\begin{figure}[htp]
\centering
\includegraphics[width=\linewidth]{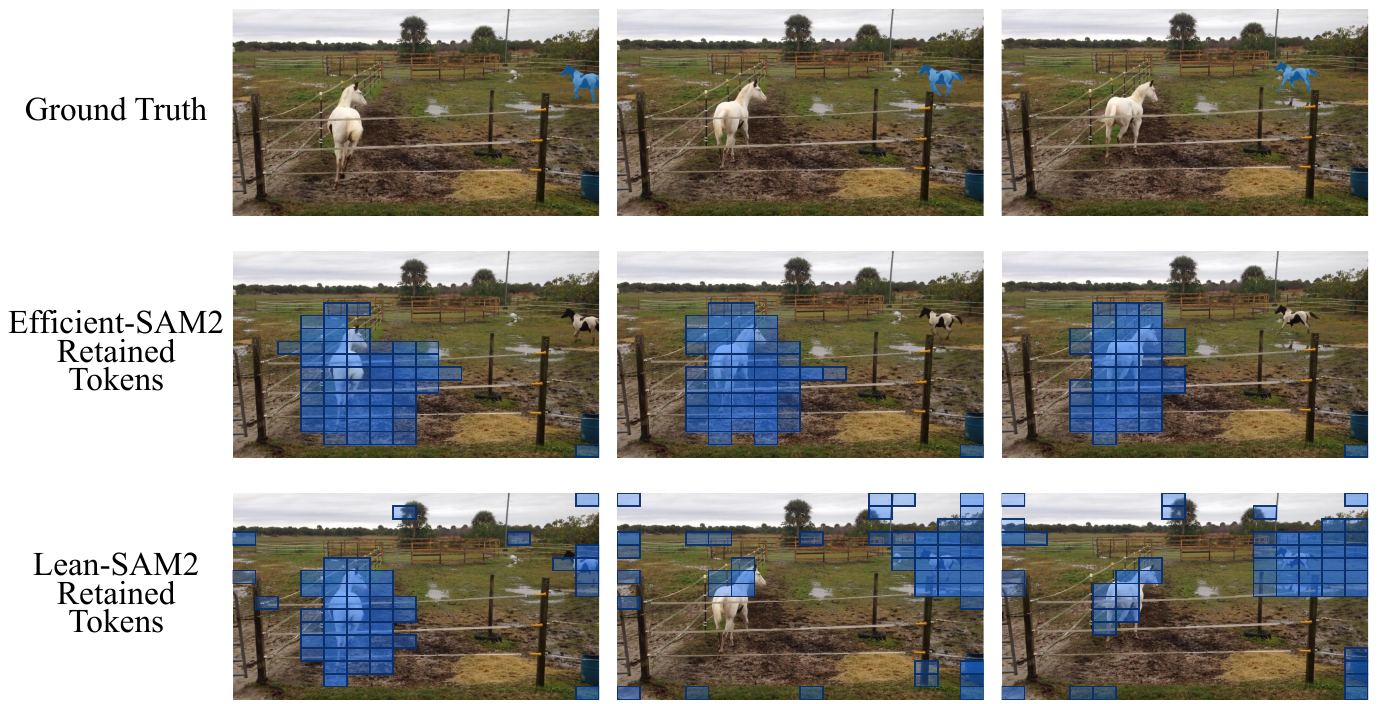}
\caption{In frames with distractors and small target regions, attention-only pruning in Efficient-SAM2 tends to retain tokens around the distractor while suppressing true target regions. In contrast, TAMP used in our Lean-SAM2 preserves target-related tokens. Frames are sourced from the LVOSv2 valid dataset using the SAM2.1-Base+ model.
}
\label{fig:tamp-corrupted-attention}
\end{figure}

\subsection{Target-Anchored Memory Pruning}
\label{sec:method-anchor-pruning}

As shown in the previous section, memory attention poses a severe computational bottleneck in VOS, primarily due to the heavy cross-attention computation between the current frame and the growing historical entries in the memory bank. To mitigate this burden, existing memory pruning mechanisms for SAM2~\cite{zhang2026efficient} typically retain memory tokens solely based on their \textit{memory attention significance}, which quantifies how strongly the current frame attends to each cached memory token. 
Specifically, given the current frame $I_t$ and the pruned working memory queue $\hat{\mathcal{M}}_t=\{m_{t-1}, \hat{m}_{t-2}, \dots, \hat{m}_{t-l}\}$, where $m_{t-1}$ is the most recent unpruned memory entry and $\{\hat{m}_{t-2}, \dots, \hat{m}_{t-l}\}$ are pruned entries, the cross-attention weight matrix $A_{t\rightarrow{t-1}} \in \mathbb{R}^{N_Q \times (H \times W) }$ between $I_t$ and $m_{t-1}$ is computed via Eq.~\ref{eq:prelim-qkv}. Here, $N_Q$ is the number of query tokens and $(H \times W)$ denotes the token count of $m_{t-1}$.
The attention significance of the $p$-th token in $m_{t-1}$ is computed by averaging its attention weights across all query positions:
\begin{equation}
s^{\text{ma}}_{t-1}[p] = \frac{1}{N_Q} \sum_{j=1}^{N_Q} A_{t\rightarrow{t-1}}[j,p], \quad p=1,\dots, HW.
\label{eq:rap-attn-score-v3}
\end{equation}

Subsequently, only a fixed subset of tokens exhibiting the highest attention significance are preserved. 
While this attention-only memory pruning mechanism is effective in clean scenarios with continuous target visibility, it frequently breaks down in complex video sequences characterized by occlusions or visually similar distractors. 
As illustrated in Fig.~\ref{fig:tamp-corrupted-attention},the attention-only pruning mechanism is susceptible to retaining tokens near distractors at the expense of true target regions, which risks permanently discarding vital target-related tokens and causing irreversible errors.

To overcome this limitation, we propose Target-Anchored Memory Pruning (TAMP), an intra-frame pruning mechanism that exploits the initial prompt memory entry to anchor critical memory tokens to the reliable target object identity. Specifically, a set of stable reference anchors is sampled from the prompt memory tokens situated within the initial object foreground, denoted as $\mathcal{M}^{\mathrm{prompt}}_{\mathrm{fg}}$. To prevent over-sampling large objects or under-sampling small ones, TAMP adaptively determines the anchor budget $K_{\text{anchor}}$ as follows:
\begin{equation}
K_{\text{anchor}} = \mathrm{clip} \left( \lceil r_{\text{anchor}} |\mathcal{M}^{\mathrm{prompt}}_{\mathrm{fg}}|  \rfloor, K_{\min}, K_{\max} \right),
\end{equation}
where $r_{\text{anchor}}$ is a predefined sampling ratio, while $K_{\min}=8$ and $K_{\max}=64$ bound the range of $K_{\text{anchor}}$.
Based on this budget, TAMP employs a diversity-driven greedy strategy to select representative anchors. At $t=1$, the initial significance scores of the foreground tokens within $\mathcal{M}^{\mathrm{prompt}}_{\mathrm{fg}}$ are computed via Eq.~\ref{eq:rap-attn-score-v3}, and the token with the highest significance is selected as the seed anchor $a_1$. The remaining anchors are then selected iteratively. For $k=2,\ldots,K_{\text{anchor}}$, TAMP greedily appends the token $a_k$ by searching for a candidate token $z$ that minimizes its maximum cosine similarity to any already selected anchor $a$ in the existing set $\mathcal{A}_{k-1}=\{a_1,\dots,a_{k-1}\}$:
%
\begin{equation}
a_k = \arg\min_{z \in \mathcal{M}^{\mathrm{prompt}}_{\mathrm{fg}} \setminus \mathcal{A}_{k-1}} \max_{a \in \mathcal{A}_{k-1}} \operatorname{CosSim}(z,a),
\end{equation}
where $\operatorname{CosSim}(\cdot,\cdot)$ denotes the cosine similarity function. The final anchor set is constructed as $\mathcal{A}=\{a_k\}_{k=1}^{K_{\text{anchor}}}$. This diversity-driven selection prevents the anchors from collapsing into a localized, redundant cluster, providing a compact yet spatially distributed representation of the initial target identity.

Afterward, for timestep $t$, TAMP evaluates the reliability of incoming memory tokens within $m_{t-1}$ by measuring their semantic consistency with the trusted anchors in $\mathcal{A}$. For the $p$-th memory token, its anchor-similarity score is defined as:
\begin{equation}
    s^{\mathrm{anc}}_{t-1}[p] = \max_{a \in \mathcal{A}} \operatorname{CosSim}\left(m_{t-1}[p], a\right), \quad p=1,\dots, HW.
\label{eq:anchor-similarity}
\end{equation}

Consequently, $s^{\mathrm{anc}}_{t-1}[p]$ reflects the semantic correlation between the memory token and the original target, ensuring that informative tokens remain protected even when the attention significance is temporarily distracted.

The final selection score for the $p$-th memory token is formulated as:
\begin{equation}
\begin{aligned}
    s_{t-1}[p] & = \frac{s^{ma}_{t-1}[p]}{\max_k s^{ma}_{t-1}[k]} \left( 1 + \lambda_\text{A}\frac{s_{t-1}^{\mathrm{anc}}[p]+1}{2} \right), 
    \\
     \quad p &=1,\dots,HW,
\end{aligned}
\end{equation}
where the first term normalizes the raw attention significance to $[0,1]$, and $\lambda_\text{A}$ is a hyperparameter controlling the strength of the anchor-guided modulation. Under this formulation, tokens that are semantically incompatible with the anchors ($s^{\mathrm{anc}}_{t-1}[p] \approx -1$) receive no adjustment, whereas tokens that highly align with the target identity ($s^{\mathrm{anc}}_{t-1}[p] \approx 1$) receive a maximum boost of $1+\lambda_\text{A}$.

Finally, token-level pruning is performed on $m_{t-1}$ based on the unified score $s_{t-1}$:
\begin{equation}
\hat{m}_{t-1} = \operatorname{TopK}\big( m_{t-1}, s_{t-1}, K_{\text{ret}} \big),
\label{eq:memory-prune}
\end{equation}
where $\operatorname{TopK}$ selects the top-$K_{\text{ret}}$ scoring tokens within $m_{t-1}$ to be retained. 
By executing TAMP at timestep $t$ to prune $m_{t-1}$ before appending the newly generated entry $m_t$, we maintain a highly reliable, intra-frame compressed working memory queue $\hat{\mathcal{M}}_{t+1}=\{m_{t}, \hat{m}_{t-1}, \hat{m}_{t-2}, \dots, \hat{m}_{t-l+1}\}$ available for the next timestep $t+1$.
Through balancing dynamic attention significance with static anchor consistency, TAMP effectively overcomes deceptive attention. 
As shown in Fig.~\ref{fig:tamp-corrupted-attention}, by leveraging TAMP, our Lean-SAM2 successfully preserves target-related tokens. This effectively mitigates the risk of irreversibly discarding vital target-related tokens and helps the model accurately identify the correct target, thereby ensuring robust intra-frame pruning against visual interference.

\begin{figure}[thp]
\centering
\includegraphics[width=\linewidth]{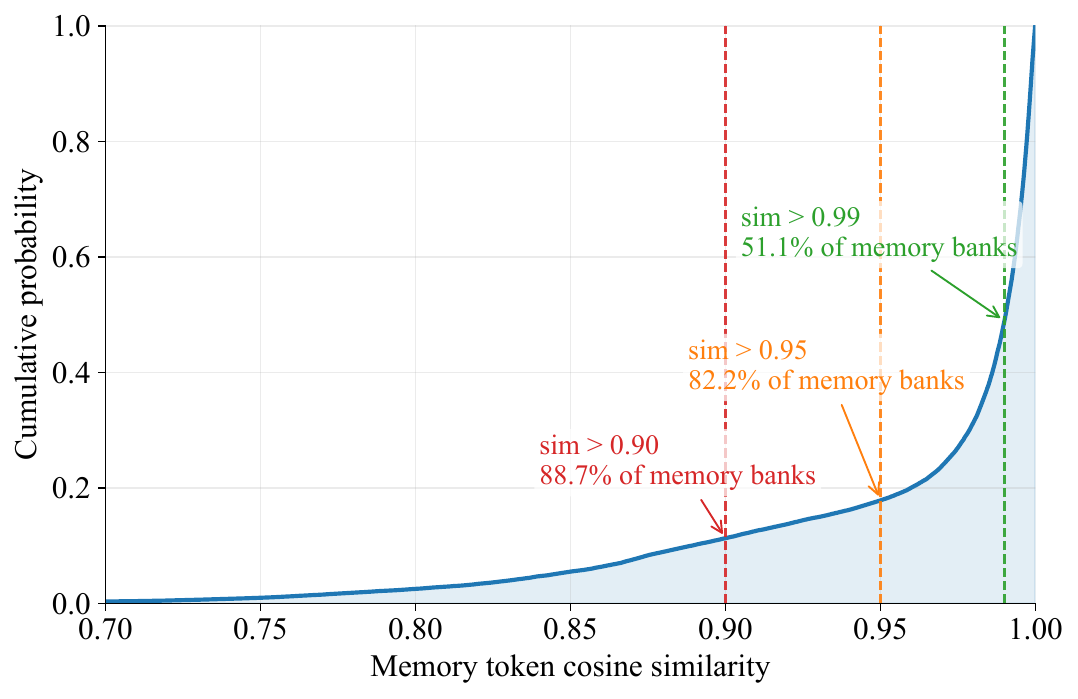}
\caption{Cumulative distribution function (CDF) of token similarity within the memory bank after TAMP. The curve demonstrates that the pruned memory tokens maintain a high degree of similarity, yielding a mean score of 0.968 and a median of 0.990. Results are reported by randomly selecting 20 videos from the LVOSv2 valid dataset using the SAM2.1-Base+ model.}
\label{fig:old-tail-similarity}
\end{figure}

\subsection{Temporal Condensation with Insurance Memory}
\label{sec:method-TCIM}

After mitigating intra-frame spatial redundancy via TAMP, the retained memory tokens across different frames predominantly capture target-related features, which inherently introduces inter-frame redundancy. As illustrated in Fig.~\ref{fig:old-tail-similarity}, the cross-frame tokens exhibit high cosine similarity. Consequently, performing cross-attention over the entire historical sequence incurs unnecessary computational overhead.

Motivated by this, we propose Temporal Condensation with Insurance Memory (TCIM), an inter-frame condensation mechanism designed to compress redundant historical memory entries while preserving vital long-term context. Formally, at time step $t+1$, TCIM restricts the working memory queue $\hat{\mathcal{M}}'_{t+1}$ to a fixed capacity of three entries:
\begin{equation}
\hat{\mathcal{M}}'_{t+1} = \{m_{t}, \hat{m}_{t-1}, \hat{H}_{t-2}\},
\end{equation}
where $m_{t}$ and $\hat{m}_{t-1}$ denote the latest unpruned memory entry and the pruned memory entry, respectively, while $\hat{H}_{t-2}$ represents the accumulated historical token summarizing all preceding context from the initial frame up to step $t-2$.
During the update, $m_{t}$ is preserved intact to maintain the most recent temporal context. 
To prevent unreliable features from corrupting the historical context during occlusions, the integration of $\hat{m}_{t-1}$ is adaptively modulated by its target visibility. Since a higher occlusion score $o_{t-1}$ indicates that the target is more visible and less obstructed, it implies higher quality. 
Grounded in this rationale, $\hat{m}_{t-1}$ is adaptively integrated with the mean of $\hat{H}_{t-2}$ via a target visibility-gated fusion to produce the updated historical token $\hat{H}_{t-1}$:
\begin{equation}
    \begin{aligned}
      \hat{H}_{t-1}[p] &= \alpha \hat{m}_{t-1}[p] + (1 - \alpha )\text{mean}(\hat{H}_{t-2}), \\
      \alpha &= \rho o_{t-1} \frac{
     \sigma\!\left(\frac{o_{t-1}}{T}\right)
  }{
    \sigma\!\left(\frac{1}{T}\right)
  },
      \label{eq:rap-TCC-average-v3}
    \end{aligned}
\end{equation}
where $p = 1, \dots, K_{\text{ret}}$ indexes the tokens, $\rho$ is a balancing hyperparameter, $T$ is the temperature, and $o_{t-1} \in [0,1] $ is the occlusion score ($o_{t-1}=1$ denotes full visibility) predicted via Eq.~\ref{eq:prelim-decoder}. 
Under this formulation, the gating coefficient $\alpha$ scales monotonically with target visibility. Specifically, a high visibility score dynamically amplifies $\alpha$ to inject reliable features into the history, whereas severe occlusions suppress $\alpha$ to safeguard the accumulated memory from unreliable segmentation results.
%
%
Following this fusion step, the memory entries temporarily condense to $\{m_{t}, \hat{H}_{t-1}\}$. 
Once the frame $I_{t+1}$ is processed, $m_{t}$ is pruned via TAMP into $\hat{m}_t$, while the newly generated entry $m_{t+1}$ is appended to the front of the queue. This yields the updated state for the next timestep $t+2$: $\hat{\mathcal{M}}'_{t+2} = \{m_{t+1}, \hat{m}_t, \hat{H}_{t-1}\}$. If fewer than three preceding frames are available, this operation is naturally bypassed.

Moreover, to alleviate the loss of critical historical memory entries, TCIM further maintains a compact insurance bank $\mathcal{M}^{\mathrm{ins}}_{t+1}$ in parallel with the compressed working memory queue $\hat{\mathcal{M}}'_{t+1}$. 
Crucially, we employ the predicted occlusion score $o_t$ as the indicator of feature quality and archive entries with high $o_t$ to serve as reliable and critical recovery anchors. 
When a new memory entry $\hat{m}_{t}$ is generated by TAMP at timestep $t+1$, TCIM conditionally enqueues it into the insurance bank $\mathcal{M}^{\mathrm{ins}}_{t+1}$ based on its occlusion score $o_t$:
\begin{equation}
\mathcal{M}^{\mathrm{ins}}_{t+1}
\begin{cases}
\operatorname{FIFO}_{B}\left(
\mathcal{M}^{\mathrm{ins}}_{t}\oplus{\hat{m}_{t}}
\right), 
& o_{t}>\theta_{\mathrm{ins}},\\
\mathcal{M}^{\mathrm{ins}}_{t}, & o_{t}\le\theta_{\mathrm{ins}},\end{cases}
\label{eq:rap-insurance-update}
\end{equation}
where $B$ denotes the maximum capacity of the insurance bank and $\theta_{\mathrm{ins}}$ is the threshold.
By applying TCIM, we obtain a highly compact but robust inter-frame compressed working memory queue including $\hat{\mathcal{M}}'_{t+1}$ and $\mathcal{M}^{\mathrm{ins}}_{t+1}$.
Consequently, the final aggregated memory bank $\mathcal{M}^{\mathrm{final}}_{t+1}$ forwarded to the memory cross-attention module is constructed by combining all three complementary components:
\begin{equation}
  \mathcal{M}^{\mathrm{final}}_{t+1} = \mathcal{M}^{\mathrm{prompt}} \oplus \hat{\mathcal{M}}'_{t+1} \oplus \mathcal{M}^{\mathrm{ins}}_{t+1}.
\label{eq:final-memory-bank}
\end{equation}

As a result, TCIM reduces inter-frame temporal redundancy significantly.
By seamlessly integrating TAMP and TCIM, our framework achieves a favorable balance between computational efficiency and segmentation accuracy, yielding a highly compact yet informative memory representation. As demonstrated in Sec.~\ref{sec:Main Performance}, compared to existing methods, our method substantially reduces computational overhead while maintaining superior segmentation accuracy.

\begin{figure}[htpt]
\centering
\includegraphics[width=\columnwidth]{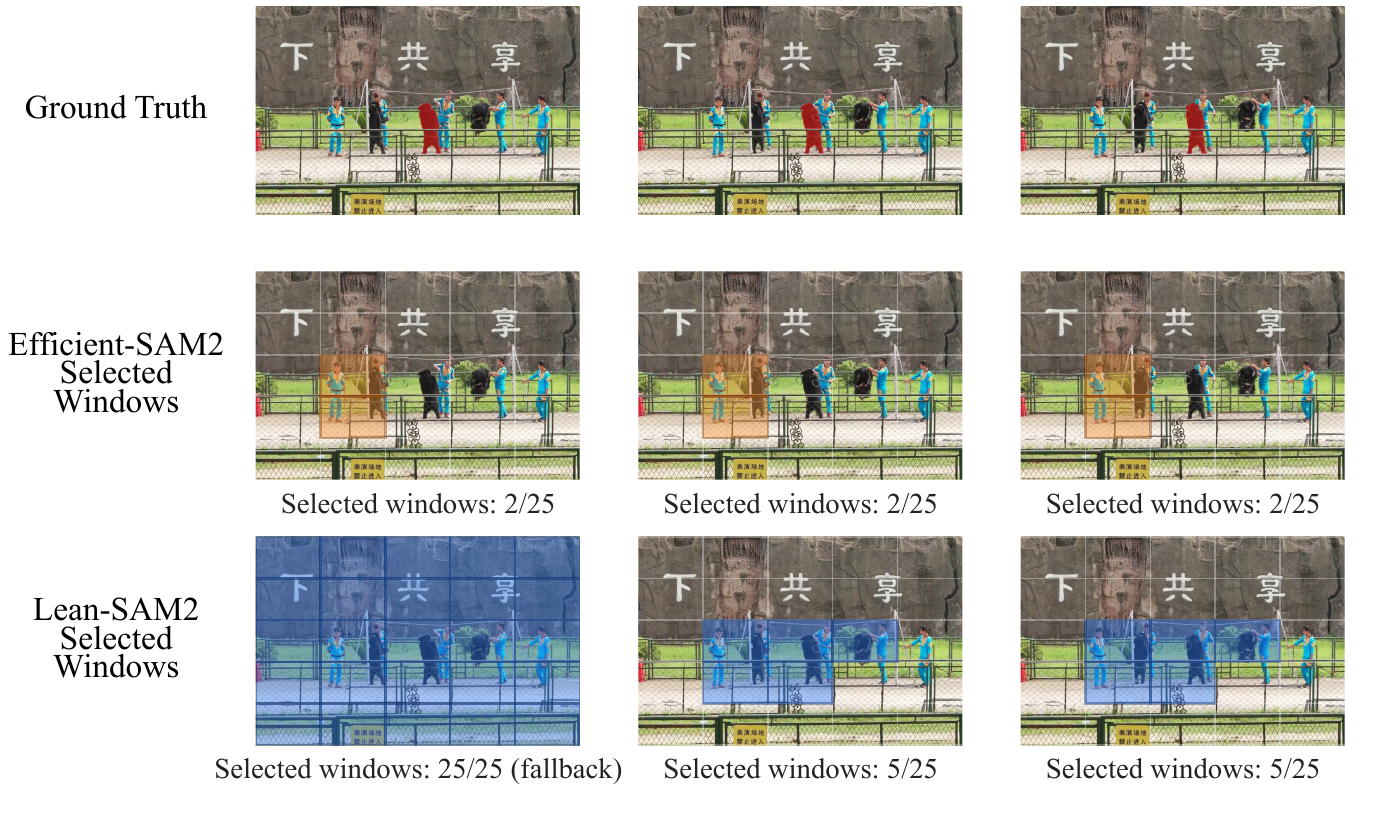}
\caption{In this frame, the predicted mask of Efficient-SAM2 drifts toward a distractor, causing it to route erroneous windows to the image encoder. In contrast, by leveraging TARR, Lean-SAM2 selects more critical or even all windows to provide sufficient contextual information. Frames are sourced from the LVOSv2 valid dataset using the SAM2.1-Base+ model.
}
\label{fig:case-study-routing}
\end{figure}

\subsection{Target-Anchored Risk-Aware Routing}
\label{sec:method-TARR}

The heavy computational footprint of the image encoder also poses a critical inference bottleneck in VOS, as illustrated in Fig.~\ref{fig:small-tiny-speed-runtime-breakdown}. To mitigate this burden, the existing method~\cite{zhang2026efficient} employs a Sparse Window Routing (SWR) strategy. It partitions each incoming frame $I_t$ into two disjoint sets of spatial windows, namely a target-relevant set $\mathcal{W}^{\mathrm{full}}_t$ processed by the heavy encoder and a background set $\mathcal{W}^{\mathrm{bypass}}_t$ routed through a lightweight, trainable bypass network. The routing decision for the current frame $I_t$ is predicted based on the states of the preceding frame $I_{t-1}$. Specifically, a window in $I_t$ is assigned to $\mathcal{W}^{\mathrm{full}}_t$ if its co-located region in $I_{t-1}$ is either covered by the predicted foreground mask or exhibits high attention significance within the memory attention module, and is otherwise relegated to the bypass branch.
However, as shown in Fig.~\ref{fig:case-study-routing}, once the model starts following the incorrect distractors, this routing strategy continuously selects windows associated with the wrong object to the image encoder, thereby reinforcing the erroneous foreground cues and further aggravating the segmentation drift.

Motivated by this, to safeguard sparse routing against prediction drift while sustaining architectural efficiency, we propose Target-Anchored Risk-Aware Routing (TARR). Building upon~\cite{zhang2026efficient}, TARR introduces a prompt-anchor evaluation mechanism and a risk-aware fallback policy to rectify erroneous window selection.
%
In particular, TARR first reformulates the window selection policy by introducing a prompt-anchor evaluation mechanism to safeguard target-bearing regions. Following the core idea of TAMP, the established anchor set $\mathcal{A}=\{a_k\}_{k=1}^{K_{\text{anchor}}}$ serves as an oracle reference of the initial target identity. Upon processing frame $I_{t-1}$, we evaluate the semantic alignment between each token in the newly generated memory entry $m_{t-1}$ and these trusted anchors, yielding the alignment map $\alpha_{t-1}$:
\begin{equation}
\alpha_{t-1}[p] = \max_{a_k \in \mathcal{A}} \operatorname{CosSim} \left( m_{t-1}[p], a_k \right), \quad p=1,\dots,HW,
\label{eq:rap-awr-anchor-affinity}
\end{equation}
where $\operatorname{CosSim}(\cdot, \cdot)$ denotes the cosine similarity and $HW$ represents the spatial resolution.

The alignment map $\alpha_{t-1}$ is subsequently utilized to guide the routing pass for the subsequent frame $I_{t}$. To achieve this, $\alpha_{t-1}$ is spatially reshaped to match the dimensions of the window grid. Treating each window $w$ as a collection of grid positions, TARR explicitly routes windows whose maximum alignment scores exceed $0.5$ to the heavy encoder via:
\begin{equation}
\mathcal{W}^{\mathrm{full}}_{t} = \mathcal{W}^{\mathrm{full}}_{t} \cup \left\{w : \max_{p\in w}\alpha_{t-1}[p] \ge 0.5\right\}.
\label{eq:rap-awr-anchor}
\end{equation}

Consequently, even if mask predictions drift or become corrupted by distractors, any spatial window capturing features semantically aligned with the initial target template will be actively processed by the heavy encoder, thereby preventing the erroneous bypass of target regions.

Furthermore, TARR implements a risk-aware fallback policy that completely overrides the sparse routing scheme whenever the segmentation state enters a volatile scenario. Specifically, if the segmentation state in the preceding frame $I_{t-1}$ exhibits either an abrupt drop in confidence score or an abnormal geometric deformation, the model temporarily abandons the bypass branch and reverts to full-capacity processing. This triggers a global frame refresh for the next frame $I_t$ by routing all spatial windows through the heavy encoder:
\begin{equation}
\mathcal{W}^{\mathrm{full}}_{t} = \mathcal{W}^{all}_t, \quad \text{if } o_{t-1} < \theta_{\text{full}} \vee \left|\frac{|O_{t-1}|}{|O_{t-2}|}-1\right| > 0.5,
\label{eq:rap-awr-refresh}
\end{equation}

where $\theta_{\text{full}}$ is the threshold, $\mathcal{W}^{all}_t$ is the set of all windows within frame $I_i$, $o_{i}$ and $|O_i|$ denote the occlusion score of frame $I_{t-1}$ and the mask area of frame $I_i$, respectively. Thus, when occlusions and distractors emerge, this fallback mechanism triggers a full-frame reactivation of the heavy encoder.
As shown in Fig.~\ref{fig:case-study-routing}, TARR routes more critical and even the entire frame to the heavy encoder when the target object is confronted with distractors, thereby providing sufficient contextual information to guarantee superior performance. Notably, this fallback mechanism is triggered at a low fallback rate of only 25\%, resulting in minimal overhead and comparable speedup to \cite{zhang2026efficient}, as discussed in Sec.~\ref{sec:Main Performance}.

\begin{table*}[t]
\centering
\definecolor{maRow}{HTML}{EAF3FF}
\definecolor{ieRow}{HTML}{FFF3E0}
\definecolor{fullRow}{HTML}{EAF7EA}
\caption{Main accuracy and efficiency comparison across backbone sizes. Scores are \(\mathcal{J}\&\mathcal{F}\) (\%).  Blue and orange rows denote the evaluation of the memory attention and image encoder modules, respectively, with the speedup reported on a per-module. Green rows represent full-pipeline methods, with the speedup reported on the overall model.}
\label{tab:result-samlarge}
\begin{tabular}{cccccccc|c}
\toprule
\textbf{Model} & \textbf{Method} & \textbf{LVOSv2 valid} & \textbf{SA-V valid} & \textbf{SA-V test} & \textbf{MOSEv2} & \textbf{MOSEv1} & \textbf{AVG.} &   \textbf{Speedup} \\
\midrule
 & Baseline & 84.2 & 78.5 & 79.8 & 49.4 & 74.6 & 73.3 &   -- \\
\cmidrule(lr){2-9}
\rowcolor{maRow}\cellcolor{white} & MemPool & 78.5 & 73.2 & 72.6 & 43.0 & 71.3 & 67.7 &  \(1.676\times\) \\
\rowcolor{maRow}\cellcolor{white} & Efficient-SAM2-SMR & 83.2 & 77.7 & \textbf{79.6} & 48.7 & 74.3 & 72.7 &   \(1.530\times\) \\
\rowcolor{maRow}\cellcolor{white} & \textbf{TAMP+TCIM} & \textbf{84.0} & \textbf{79.9} & \textbf{79.6}  & \textbf{48.9} & \textbf{74.4} & \textbf{73.4} &  \textbf{\(1.679\times\)} \\
\cmidrule(lr){2-9}
\rowcolor{ieRow}\cellcolor{white} & ToMe & 49.4 & 72.4 & 47.8 & 28.2 & 44.7 & 48.5 & \(1.221\times\) \\
\rowcolor{ieRow}\cellcolor{white} & ALGM & 57.0 & 53.5 & 54.9 & 31.0 & 67.0 & 52.7 & \(1.352\times\) \\
\rowcolor{ieRow}\cellcolor{white} & ToMe4DM & 76.9 & 69.6 & 71.7 & 41.5 & 66.1 & 65.2 &  \(1.080\times\) \\
\rowcolor{ieRow}\cellcolor{white} & Efficient-SAM2-SWR & 79.0 & 76.9 & 78.6 & 45.1 & 73.3 & 70.6 &   \textbf{\(1.376\times\)} \\
\rowcolor{ieRow}\cellcolor{white} & \textbf{TARR} & \textbf{82.6} & \textbf{78.5} & \textbf{79.9} & \textbf{48.9} & \textbf{74.6} & \textbf{72.9} &   \(1.364\times\) \\
\cmidrule(lr){2-9}
\rowcolor{fullRow}\cellcolor{white} & Efficient-SAM2-full & 78.6 & 76.3 & 78.8 & 45.2 & 72.6 & 70.3 &  \(1.311\times\) \\
\rowcolor{fullRow}\cellcolor{white}\multirow{-11}{*}{SAM2.1-Large} & \textbf{Lean-SAM2} & \textbf{83.6} & \textbf{79.8} & \textbf{79.5} & \textbf{48.5} & \textbf{74.3} & \textbf{73.1} & \textbf{\(1.412\times\)} \\
\midrule
 & Baseline & 83.6 & 77.6 & 78.1 & 43.5 & 73.5 & 71.3 &  -- \\
\cmidrule(lr){2-9}
\rowcolor{maRow}\cellcolor{white} & MemPool & 77.8 & 72.8 & 72.3 & 41.2 & 71.0 & 67.0 &   \textbf{\(1.761\times\)} \\
\rowcolor{maRow}\cellcolor{white} & Efficient-SAM2-SMR & 81.1 & 76.8 & 77.5 & 43.6 & \textbf{73.2} & 70.4 &   \(1.576\times\) \\
\rowcolor{maRow}\cellcolor{white} & \textbf{TAMP+TCIM} & \textbf{83.0} & \textbf{78.2} & \textbf{78.1} & \textbf{45.0} & 72.3 & \textbf{71.3} & \(1.610\times\) \\
\cmidrule(lr){2-9}
\rowcolor{ieRow}\cellcolor{white} & ToMe & 61.9 & 55.0 & 55.3 & 41.2 & 53.5 & 53.4 &  \(1.170\times\) \\
\rowcolor{ieRow}\cellcolor{white} & ALGM & 78.6 & 70.8 & 71.9 & 42.3 & 67.0 & 66.1 &  \(1.027\times\) \\
\rowcolor{ieRow}\cellcolor{white} & ToMe4DM & 70.3 & 63.8 & 62.7 & 37.2 & 59.5 & 58.7 &  \(1.000\times\) \\
\rowcolor{ieRow}\cellcolor{white} & Efficient-SAM2-SWR & 78.3 & 74.3 & 75.2 & 42.3 & 71.2 & 68.3 &  \(1.290\times\) \\
\rowcolor{ieRow}\cellcolor{white} & \textbf{TARR} & \textbf{81.4} & \textbf{77.1} & \textbf{77.5} & \textbf{43.6} & \textbf{73.2} & \textbf{70.6} &  \textbf{\(1.351\times\)} \\
\cmidrule(lr){2-9}
\rowcolor{fullRow}\cellcolor{white} & Efficient-SAM2-full & 78.8 & 74.3 & 75.0 & 42.1 & 70.9 & 68.2  & \(1.254\times\) \\
\rowcolor{fullRow}\cellcolor{white}\multirow{-11}{*}{SAM2.1-Base+} & \textbf{Lean-SAM2} & \textbf{82.4} & \textbf{78.3} & \textbf{77.4} & \textbf{44.2} & \textbf{71.6} & \textbf{70.8}  & \textbf{\(1.417\times\)} \\
\midrule
 & Baseline & 81.7 & 76.9 & 76.8 & 47.6 & 71.1 & 70.8 &  -- \\
\cmidrule(lr){2-9}
\rowcolor{maRow}\cellcolor{white} & MemPool & 76.1 & 69.1 & 69.1 & 39.3 & 65.8 & 63.9 &      \textbf{\(1.912\times\)} \\
\rowcolor{maRow}\cellcolor{white} & Efficient-SAM2-SMR & 82.7 & 77.0 & 76.7 & 46.2 & \textbf{72.5} & 71.0 &   \(1.676\times\) \\
\rowcolor{maRow}\cellcolor{white} & \textbf{TAMP+TCIM} & \textbf{83.1} & \textbf{78.5} & \textbf{79.1} & \textbf{46.4} & 71.9 & \textbf{71.8} &  \(1.662\times\) \\
\cmidrule(lr){2-9}
\rowcolor{ieRow}\cellcolor{white} & ToMe & 82.1 & 73.8 & 73.9 & 45.4 & 70.3 & 69.1 &     \(1.179\times\) \\
\rowcolor{ieRow}\cellcolor{white} & ALGM & 82.0 & 72.5 & 73.7 & 44.9 & 67.8 & 68.2 &     \(1.028\times\) \\
\rowcolor{ieRow}\cellcolor{white} & ToMe4DM & 73.6 & 64.0 & 63.8 & 37.8 & 58.5 & 59.5 &     \(1.000\times\) \\
\rowcolor{ieRow}\cellcolor{white} & Efficient-SAM2-SWR & 80.7 & 74.5 & 76.0 & 45.6 & 70.7 & 69.5 &   \(1.306\times\) \\
\rowcolor{ieRow}\cellcolor{white} & \textbf{TARR} & \textbf{82.5} & \textbf{75.9} & \textbf{76.7} & \textbf{46.9} & \textbf{72.6} & \textbf{70.9} &   \textbf{\(1.313\times\)} \\
\cmidrule(lr){2-9}
\rowcolor{fullRow}\cellcolor{white} & Efficient-SAM2-full & 79.6 & 75.4 & 75.6 & 45.3 & 70.4 & 69.3  & \(1.296\times\) \\
\rowcolor{fullRow}\cellcolor{white}\multirow{-11}{*}{SAM2.1-Small} & \textbf{Lean-SAM2} & \textbf{81.6} & \textbf{77.6} & \textbf{79.0} & \textbf{46.3} & \textbf{71.1} & \textbf{71.1} &  \textbf{\(1.433\times\)} \\

\bottomrule
\end{tabular}%
\end{table*}

\section{Experiments}
\label{sec:experiments}

\providecommand{\better}[1]{\textbf{#1}}
\providecommand{\oursstar}{$\star$}

\subsection{Experimental Setup}
\label{sec:exp-setup}

\textbf{Datasets and evaluation metrics.} We evaluate the performance of Lean-SAM2 across various VOS benchmarks, including the validation sets of LVOSv2, MOSEv2, MOSEv1, and SA-V, and the test set of SA-V~\cite{hong2025lvos,ding2025mosev2,ding2023mose,ravi2025sam}. Notably, the LVOSv2 dataset is a challenging benchmark that assesses long-term memory robustness, distractor handling, and target recovery capability under severe occlusions. Following standard evaluation protocols, we report the $\mathcal{J}\&\mathcal{F}$ score.

\textbf{Timing protocol.} For a fair evaluation, we conduct all speedup experiments on 20 video sequences from the LVOSv2 dataset and measure inference latency on a single NVIDIA RTX 3090 GPU. The speedup is reported relative to the official SAM2.1 baseline.

\textbf{Baseline and competing methods.} We compare Lean-SAM2 with the state-of-the-art acceleration frameworks, focusing on the recent Efficient-SAM2~\cite{zhang2026efficient}, alongside other representative post-training acceleration methods in vision transformer including ToMe~\cite{bolya2022token}, ALGM~\cite{norouzi2024algm}, and ToMe4DM\cite{bolya2023token}.
To provide a comprehensive analysis, we also evaluate individual modules of Efficient-SAM2, where Efficient-SAM2-SMR and Efficient-SAM2-SWR denote the variants adopting only Sparse Memory Retrieval and Sparse Window Routing, respectively, while Efficient-SAM2-Full represents the unified framework combining both.

\textbf{Implementation Details.}
Notably, TAMP and TCIM are training-free mechanisms, whereas the shortcut branch employed in TARR maintains the identical architecture and training configurations as \cite{zhang2026efficient} to ensure a fair comparison. 
Specifically, the shortcut branch is trained on 30 videos selected from the SA-V training set using the AdamW optimizer with a learning rate of $1\times10^{-4}$ for 3 epochs. The sampling stride is set to 3 for the Small, Base+ variants, and 4 for the Large variant, with the update stride fixed at 32. 
The hyperparameters $r_{\text{anchor}}$ and $\lambda_\text{A}$ used in TAMP are set to $0.05$ and $2.0$, respectively. For TCIM, the parameters $\rho$, $\theta_{\text{ins}}$, $T$, and the size of $\mathcal{M}^{\text{ins}}$ are set as $0.55$, $0.7$, $0.2$, and $3$, respectively. For TARR, the $\theta_{\text{full}}$ is set to 0.99.
All codes are implemented with Pytorch and executed on a single NVIDIA RTX 3090 GPU.

\textbf{Timing protocol.} All speed evaluations are conducted on a single NVIDIA RTX 3090 GPU using the same 20 videos sampled from the LVOSv2 valid dataset. Following \cite{zhang2026efficient}, we benchmark all methods at Float32 precision, and the speedup is reported relative to the official SAM2.1 baseline evaluated under the same setting.


\subsection{Main Performance}
\label{sec:Main Performance}

\textbf{SAM2.1-Large.} The results on SAM2.1-Large are reported in the top section of Tab.~\ref{tab:result-samlarge}. When accelerating on the memory attention module, the proposed TAMP+TCIM demonstrates a better performance and efficiency than Efficient-SAM2-SMR. For instance, TAMP+TCIM respectively achieves gains of $0.8\%$ and $2.2\%$ on the LVOSv2 valid and SA-V valid datasets while sustaining a higher module-level speedup of $1.679\times$.
For the acceleration of the image encoder module, TARR consistently outperforms existing methods. For instance, TARR achieves an $82.6\%$ $\mathcal{J}\&\mathcal{F}$ score on the LVOSv2 valid dataset and a $79.9\%$ on the SA-V test dataset, outperforming Efficient-SAM2-SWR by $3.6\%$ and $1.3\%$ respectively. At the same time, it delivers a comparable module-level speedup of $1.364\times$.
Finally, when evaluating full-model acceleration, Lean-SAM2 significantly surpasses Efficient-SAM2-full. It shows gains of $5.0\%$ on LVOSv2 valid dataset and $3.3\%$ on MOSEv2 dataset. Moreover, it accelerates the full-model inference by $1.412\times$, which substantially outperforms the $1.311\times$ speedup achieved by Efficient-SAM2-full.

\textbf{SAM2.1-Base+.} 
The results on the SAM2.1-Base+ model are presented in the middle section of Tab.~\ref{tab:result-samlarge}. When applying acceleration to the memory attention module, our proposed TAMP+TCIM presents a good balance between performance and computational savings. Specifically, TAMP+TCIM not only sustains a substantial speedup of $1.610\times$ but also consistently delivers a performance on par with or superior to the Efficient-SAM2-SMR. For instance, it achieves gains of 1.9\% and 1.4\% on the LVOSv2 valid and SA-V valid datasets, respectively.
For image encoder acceleration, our TARR consistently surpasses existing approaches while providing a higher speedup of 
1.351$\times$. For example, TARR achieves scores of 81.4\% and 77.1\% on the LVOSv2 valid and SA-V valid datasets, outperforming the best competitor by 3.1\% and 2.8\%, respectively.
Finally, when considering full-model acceleration, our Lean-SAM2 significantly outperforms Efficient-SAM2-full, including a notable 
3.6\% gain on the LVOSv2 valid dataset. Crucially, Lean-SAM2 achieves these accuracy improvements while accelerating full-model inference to 1.417$\times$, substantially surpassing the 1.254$\times$ speedup delivered by Efficient-SAM2-full.

\textbf{SAM2.1-Small.}
The results on SAM2.1-Small are reported in the bottom section of Tab.~\ref{tab:result-samlarge}. When accelerating on the memory attention module, the proposed TAMP+TCIM demonstrates a good balance between performance and computational efficiency. 
In particular, while MemPool maximizes the memory attention module speedup to $1.912\times$, it suffers from a severe performance degradation across all benchmarks. In contrast, TAMP+TCIM consistently delivers competitive or superior performance against Efficient-SAM2-SMR while sustaining a substantial module-level speedup of $1.662\times$. For instance, TAMP+TCIM achieves notable gains of $1.5\%$ and $2.4\%$ on the SA-V valid and test datasets, respectively.
When accelerating the image encoder module, TARR consistently outperforms existing methods such as ToMe, ALGM, ToMe4DM, and Efficient-SAM2-SWR while yielding a higher speedup. For instance, TARR achieves a $82.5\%$ $\mathcal{J}\&\mathcal{F}$ score on the LVOSv2 valid dataset and a $75.9\%$ on the SA-V valid dataset, outperforming the SOTA by $1.8\%$ and $1.4\%$ respectively. 
At the same time, it delivers a superior module-level speedup of $1.313\times$. 
Furthermore, when evaluating full-model acceleration, Lean-SAM2 significantly surpasses Efficient-SAM2-full. Notably, Lean-SAM2 establishes superior performance across all benchmarks, including notable gains of $2.0\%$ on the LVOSv2 valid dataset and $3.4\%$ on the SA-V test dataset. Crucially, it accomplishes these accuracy improvements while accelerating full-model inference to $1.433\times$, which substantially outperforms the $1.296\times$ speedup achieved by Efficient-SAM2-full.

In summary, the consistent performance gains and substantial speedups verified across all models and five diverse datasets demonstrate the effectiveness of the proposed Lean-SAM2.

\begin{figure}[!thpt]
\centering
\includegraphics[width=0.7\linewidth]{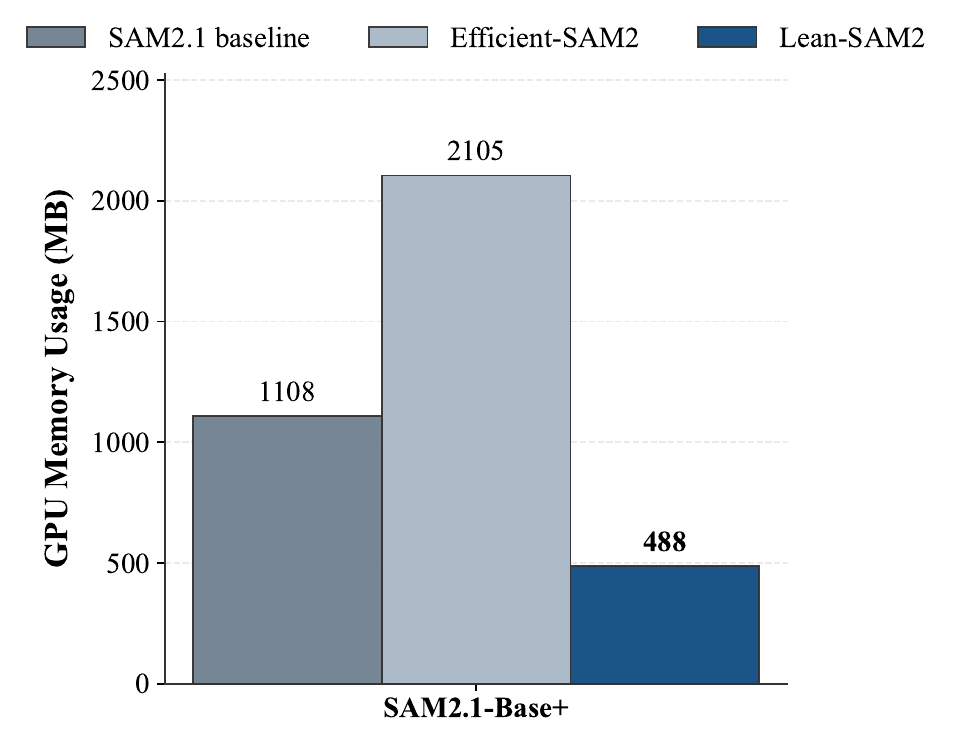}
\caption{GPU memory footprint comparison within the memory attention module.}
\label{fig:ma_memory}
\end{figure}

\subsection{Memory Reduction on Memory Attention}
\label{sec:resource-comparison}

As illustrated in Fig.~\ref{fig:ma_memory}, Lean-SAM2 significantly reduces the GPU memory footprint of the memory attention module. Specifically, Efficient-SAM2 suffers from a heavier memory overhead than the baseline due to its introduction of a layer-wise mask retrieval operation without performing real pruning on the memory bank. In contrast, Lean-SAM2 circumvents these overheads by performing a real pruning operation. It reduces the memory attention module's footprint to 488 MB, yielding a 55.9\% reduction over the baseline and a 76.8\% saving over Efficient-SAM2. These results validate the efficacy of Lean-SAM2 in mitigating computational overheads, rendering it highly viable for resource-constrained real-world deployment.

\begin{table}[ht]
\centering
\caption{Mechanism ablations on LVOSv2 valid with the SAM2.1-Base+ backbone. Speedup reports the overall model inference speedup.}
\label{tab:Mechanism-ablation}
\begin{tabular}{ccc}
\toprule
Method  & J\&F &   Speedup \\
\midrule
Baseline & 83.6 &  \(1.000\times\) \\  \hline  
+TAMP &  80.8 & \(1.281\times\)  \\
+TCIM &  84.5 & \(1.091\times\)  \\
+TAMP+TCIM & 83.0 & \(1.284\times\)  \\
+TARR & 81.4 & \(1.228\times\)    \\
TAMP+TCIM+TARR (Lean-SAM2) &  82.4 & \(1.417\times\) \\
\bottomrule
\end{tabular}
\end{table}

\begin{table*}[ht]
\centering
\caption{Ablation studies on different hyper-parameters for Lean-SAM2. We report the performance on the SAM2.1-Base+. The default choice for each parameter is highlighted in \textbf{bold}.}
\label{tab:hyper-parameter-ablation}
\begin{tabular}{ccccccccccccccccc}
\toprule
\multicolumn{2}{c}{\textbf{(a) $r_{\text{anchor}}$}} & \phantom{a} & \multicolumn{2}{c}{\textbf{(b) $\lambda_\text{A}$}} & \phantom{a} & \multicolumn{2}{c}{\textbf{(c) $\rho$}} & \phantom{a} & \multicolumn{2}{c}{\textbf{(d) $\theta_{\mathrm{ins}}$}} &  \phantom{a} & \multicolumn{2}{c}{\textbf{(e) Size of $\mathcal{M}^{\mathrm{ins}}$}} & \multicolumn{2}{c}{\textbf{(f) $T$}} \\
\cmidrule(lr){1-2} \cmidrule(lr){4-5} \cmidrule(lr){7-8} \cmidrule(lr){10-11} \cmidrule(lr){13-14} \cmidrule(lr){15-16}
Value & $\mathcal{J}\&\mathcal{F}$ & & Value & $\mathcal{J}\&\mathcal{F}$ & & Value & $\mathcal{J}\&\mathcal{F}$ & & Value & $\mathcal{J}\&\mathcal{F}$ & & Value & $\mathcal{J}\&\mathcal{F}$ &  Value & $\mathcal{J}\&\mathcal{F}$ \\
\midrule
0.01  & 82.2          & & 0 & 81.2 & & 0.45          & 81.7          & & 0.5          & 81.5          & & 0 & 76.4 & 0.05 & 81.9 \\
0.025 & 82.2          & & 1 & 82.2 & & 0.50          & 82.0          & & 0.6          & 81.8          & & 1 & 78.7 & 0.1 & 81.8 \\
\textbf{0.05} & \textbf{82.4} & & \textbf{2}  & \textbf{82.4} & & \textbf{0.55} & \textbf{82.4} & & \textbf{0.7} & \textbf{82.4} & & 2 & 80.3 &  \textbf{0.2} & \textbf{82.4}\\
0.075 & 82.3          & & 3 & 81.8 & & 0.60          & 81.4          & & 0.8          & 82.1          & & \textbf{3} & \textbf{82.4} & 0.5 & 81.7 \\
0.1   & 82.3          & & 4 & 81.7 & & 0.65          & 81.4          & & 0.9          & 81.9          & & 4 & 79.5 & 1.0 & 81.6 \\
\bottomrule
\end{tabular}
\end{table*}

\begin{table}[ht]
\centering
\caption{Ablation studies on \(\theta_{\text{full}}\). We report the performance, fallback trigger rate, and speedup on the SAM2.1-Base+. The default choice for each parameter is highlighted in \textbf{bold}. The speedup is evaluated with the overall model.}
\label{tab:theta-fallback-ablation}
\begin{tabular}{cccc}
\toprule
 \(\theta_{\text{full}}\) & \(\mathcal{J}\&\mathcal{F}\) & Fallback rate & Speedup \\
\midrule
Baseline & 83.6 & -- & \(1.000\times\)\\  \hline
 0.980 & 80.5 & 21\%  & \(1.444\times\) \\
  0.985 & 80.2 & 22\% & \(1.428\times\) \\
  \textbf{0.990} & \textbf{82.4} & \textbf{25\%} & \textbf{\(1.417\times\)} \\
 0.995 & 82.4 & 26\% & \(1.416\times\) \\
 1.000 & 83.0 & 100\% & \(1.284\times\) \\
\bottomrule
\end{tabular}
\end{table}

\begin{figure*}[ht]
    \centering

    \includegraphics[width=\textwidth]
    {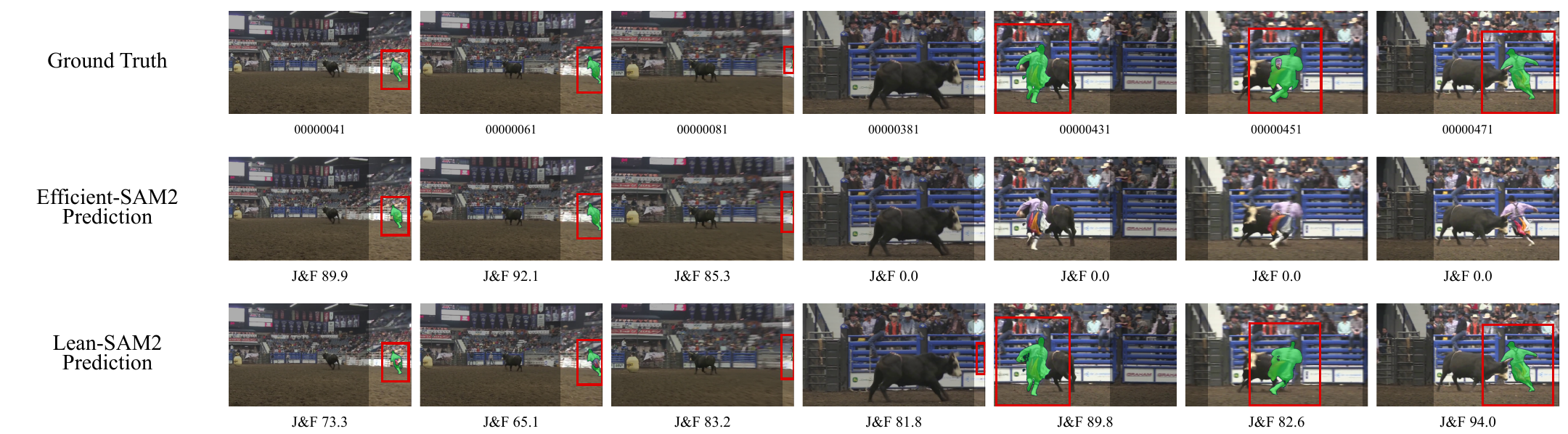}


    \includegraphics[width=\textwidth]
    {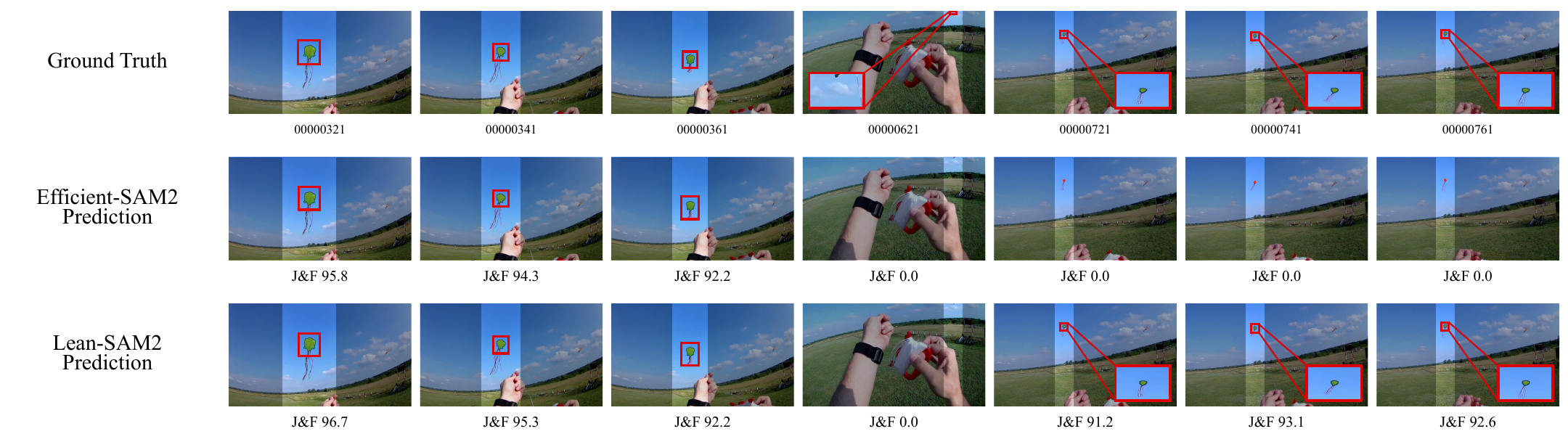}



    \caption{Qualitative comparison between Efficient-SAM 2 and Lean-SAM2, with Ground Truth provided for reference. }
    \label{fig:performance}
\end{figure*}

\subsection{Ablation Studies}
\label{sec:focused-ablation}

All ablation studies are conducted on the LVOSv2 valid dataset using the SAM 2.1-Base+ model, with computations executed on a single NVIDIA RTX 3090 GPU.

\textbf{Ablation on the proposed mechanisms.}
Tab.~\ref{tab:Mechanism-ablation} investigates the individual and joint contributions of our proposed mechanisms. Incorporating TAMP alone accelerates the baseline to $1.281\times$ but causes an accuracy drop to $80.8\%$. Conversely, integrating TCIM significantly boosts the $\mathcal{J}\&\mathcal{F}$ score to $84.5\%$ while sustaining a modest speedup of $1.091\times$, verifying its efficacy in capturing temporal relationships. 
Combining TAMP and TCIM recovers a competitive accuracy of $83.0\%$ while maintaining a high inference speedup of $1.284\times$. Furthermore, deploying TARR independently yields a $1.228\times$ speedup with a score of $81.4\%$. Ultimately, by unifying all three mechanisms, Lean-SAM2 harmonizes their individual strengths to deliver a substantial overall speedup of $1.417\times$ while sustaining a competitive $\mathcal{J}\&\mathcal{F}$ score of $82.4\%$.

\textbf{Ablation on the $r_{\text{anchor}}$.} As reported in Tab.~\ref{tab:hyper-parameter-ablation}(a), we investigate the impact of the $r_{\text{anchor}}$ by varying its value from 0.01 to 0.1. Lean-SAM2 exhibits robust performance across different values. The model achieves its optimal $\mathcal{J}\&\mathcal{F}$ score of 82.4\% at $r_{\text{anchor}} = 0.05$, which is selected as the default. Either lowering or raising this hyperparameter results in a marginal performance drop.

\textbf{Ablation on the $\lambda_\text{A}$.} We study the sensitivity of the penalty factor $\lambda_\text{A}$ within the range of 0 to 4. As shown in Tab.~\ref{tab:hyper-parameter-ablation}(b), setting $\lambda_\text{A} = 0$ drops the score to 81.2\%, highlighting the necessity of introducing the anchor-similarity score. Performance peaks at $\lambda_\text{A} = 2$ with an 82.4\% $\mathcal{J}\&\mathcal{F}$ score, which serves as our default configuration. Further increasing the penalty factor to 4 degrades the performance to 81.7\%, suggesting that a proper balance is needed between the anchor-similarity score and the attention significance.

\textbf{Ablation on the $\rho$.} We analyze the effect of the $\rho$ by scaling it from 0.45 to 0.65. As reported in Tab.~\ref{tab:hyper-parameter-ablation}(c), a small $\rho$ of 0.45 yields a suboptimal score of 81.7\%. The performance reaches 82.4\% at $\rho = 0.55$, confirming an optimal trade-off. Enlarging the ratio further up to 0.65 does not bring additional gains but rather saturates at 81.4\%.

\textbf{Ablation on the $\theta_{\mathrm{ins}}$.} We vary the confidence threshold $\theta_{\mathrm{ins}}$ from 0.5 to 0.9. Tab.~\ref{tab:hyper-parameter-ablation}(d) reveals that a loose threshold of 0.5 limits the performance to 81.5\%, whereas a strict threshold of $0.9$ similarly leads to a degraded score of 81.9\%. Lean-SAM2 reaches its peak performance of 82.4\% at the default value of 0.7, suggesting that a moderate threshold successfully selects valuable memory entries.

\textbf{Ablation on the size of the insurance bank.} We evaluate different sizes for $\mathcal{M}^{\mathrm{ins}}$. As shown in Tab.~\ref{tab:hyper-parameter-ablation}(e), disabling the insurance bank drops the $\mathcal{J}\&\mathcal{F}$ score to 76.4\%. Performance peaks at a bank size of 3, reaching 82.4\%, which we adopt as the default. Increasing the size to 4 degrades the score to 79.5\%, indicating that an oversized bank introduces redundant entries that can mislead the useful information.

\textbf{Ablation on the $T$.} 
We vary the temperature $T$ from 0.05 to 1.0. As shown in Tab.~\ref{tab:hyper-parameter-ablation}(f), a small value of 0.05 limits the performance to 81.9\%, whereas a big value of 1.0 also leads to a degraded performance of 81.6\%. The best result of 82.4\% is achieved when the $T=0.2$.

\textbf{Ablation on $\theta_{\text{full}}$}. As reported in Table~\ref{tab:theta-fallback-ablation}, a smaller $\theta_{\text{full}}$ reduces the fallback trigger rate and maximizes execution speed, albeit at the cost of tracking accuracy. 
For instance, setting $\theta_{\text{full}}=0.980$ achieves a $1.444\times$ speedup over the baseline but drops the $\mathcal{J}\&\mathcal{F}$ score to $80.5\%$. 
As $\theta_{\text{full}}$ increases, the fallback trigger rate rises monotonically. Our default configuration of $\theta_{\text{full}} = 0.990$ strikes an optimal trade-off between performance and efficiency, yielding an $82.4\%$ $\mathcal{J}\&\mathcal{F}$ score alongside a $1.417\times$ speedup at a $25\%$ trigger rate. Conversely, when $\theta_{\text{full}}$ is set to $1.000$, the fallback mechanism is activated across all frames, which is equivalent to entirely disabling the TARR module and relying solely on pure TAMP and TCIM, recovering a higher score of $83.0\%$ $\mathcal{J}\&\mathcal{F}$ but at the cost of a degraded speedup of $1.284\times$.


\begin{figure}[htpb]
\centering
\includegraphics[width=\linewidth]{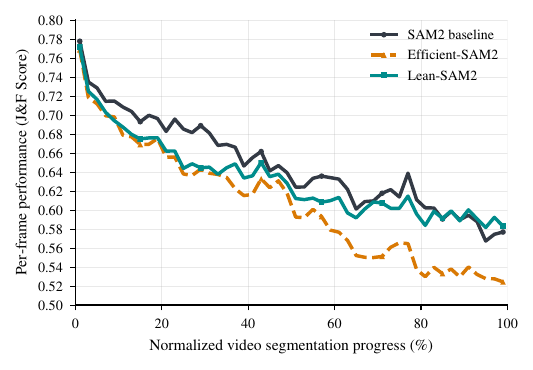}
\caption{
The per-frame $\mathcal{J}\&\mathcal{F}$ scores of the predicted masks at different timestamps, where Lean-SAM2 demonstrates stronger resilience to video segmentation compared to Efficient-SAM2. Results are reported on the full LVOSv2 valid using the SAM2.1-Base+ model.
}%
\label{fig:error-accumulation}
\end{figure}

\subsection{Per-Frame Performance Across Segmentation Progress}
\label{sec:Temporal Performance DegradationAnalysis}

Fig.~\ref{fig:error-accumulation} presents a per-frame performance analysis, plotting the $\mathcal{J}\&\mathcal{F}$ score against normalized video segmentation progress. It can be observed that the baseline method suffers from a gradual performance decay caused by error accumulation during long-term video segmentation. Notably, Efficient-SAM2 experiences a steep and continuous decline, with its score dropping below 0.54 toward the end of the progress. In contrast, Lean-SAM2 exhibits significantly higher temporal robustness against error propagation. Its performance curve closely tracks the original SAM2 baseline throughout the video and achieves slightly superior accuracy (0.58–0.60 $\mathcal{J}\&\mathcal{F}$) in the final 20\% of the timesteps, effectively alleviating temporal decay while maintaining high efficiency. This confirms that the proposed mechanisms in Lean-SAM2 preserve strong segmentation resilience when handling complex video scenarios.

Fig.~\ref{fig:performance} provides two qualitative examples demonstrating the robustness of Lean-SAM2 under challenging scenarios featured by severe occlusions and background distractors. In the upper section (rows 1–3), the target undergoes tracking difficulties due to erratic movements and complex interactions with the bull. From frame 00000381, Efficient-SAM2 completely loses track of the target, resulting in catastrophic failures across all subsequent frames. In contrast, Lean-SAM2 successfully recovers and accurately segments the target throughout the remainder of the video. A similar trend is observed in the lower section (rows 4–6). At frame 00000621, Efficient-SAM2 suffers from severe tracking drift and fails to re-localize the target even after the viewpoint stabilizes (frames 00000721 to 00000761). Conversely, Lean-SAM2 robustly re-identifies and tracks the target with high precision as soon as it reappears. These visual results qualitatively confirm that Lean-SAM2 possesses superior capabilities when handling complex scenarios replete with occlusions and distractors.

\section{Conclusion}
 
In this paper, we have presented Lean-SAM2, a holistic and lightweight framework designed to systematically eliminate the multi-level computational redundancies inherent in SAM2. 
By introducing three collaborative mechanisms including Target-Anchored Memory Pruning (TAMP), Temporal Condensation with Insurance Memory (TCIM), and Target-Anchored Risk-Aware Routing (TARR), Lean-SAM2 successfully overcomes the vulnerabilities of recent pruning and routing methods under complex scenarios containing severe occlusions and semantic distractors. Extensive experiments across multiple SAM2 variants and benchmarks consistently demonstrate that Lean-SAM2 establishes a superior balance between accuracy and efficiency, offering valuable insights for deploying SAM2 in real-time and resource-constrained applications.

\bibliographystyle{IEEEtran}
\bibliography{references}


\end{document}